# Forecasting Volcanic Radiative Power (VPR) at Fuego Volcano Using Bayesian Regularized Neural Network

Snehamoy Chatterjee
*Department of Geological and Mining Engineering and Sciences*
Michigan Technological University
Houghton, USA
schatte1@mtu.edu

Greg Waite
*Department of Geological and Mining Engineering and Sciences*
Michigan Technological University
Houghton, USA
gpwaite@mtu.edu

Sidike Paheding
*Department of Computer Science and Engineering*
Fairfield University
Fairfield, USA
spaheding@farifield.edu

Luke Bowman
*Department of Geological and Mining Engineering and Sciences*
Michigan Technological University
Houghton, USA
ljbowman@mtu.edu

*Abstract*—Forecasting volcanic activity is critical for hazard assessment and risk mitigation. Volcanic Radiative Power (VPR), derived from thermal remote sensing data, serves as an essential indicator of volcanic activity. In this study, we employ Bayesian Regularized Neural Networks (BRNN) to predict future VPR values based on historical data from Fuego Volcano, comparing its performance against Scaled Conjugate Gradient (SCG) and Levenberg-Marquardt (LM) models. The results indicate that BRNN outperforms SCG and LM, achieving the lowest mean squared error (1.77E+16) and the highest R² value (0.50), demonstrating its superior ability to capture VPR variability while minimizing overfitting. Despite these promising results, challenges remain in improving the model's predictive accuracy. Future research should focus on integrating additional geophysical parameters, such as seismic and gas emission data, to enhance forecasting precision. The findings highlight the potential of machine learning models, particularly BRNN, in advancing volcanic activity forecasting, contributing to more effective early warning systems for volcanic hazards.

I. INTRODUCTION

Volcanic eruptions can have devastating effects on human life, infrastructure, and the environment, making the study of volcanic activity and its precursors crucial for hazard assessment and risk mitigation [1]. Continuous monitoring is particularly important for volcanoes near populated areas. As magma ascends toward the surface, various geophysical and geochemical signals can be detected, including surface deformation, thermal anomalies, seismic activity, structural changes within the edifice, and variations in gas emissions—all of which contribute to forecasting eruptions [2]. Recent advancements in artificial intelligence (AI) and machine learning (ML), combined with remote sensing techniques, have substantially improved the analysis and interpretation of these complex datasets, enhancing the accuracy of eruption forecasts [3].

Remote sensing technologies, such as satellite imagery, have significantly improved our ability to monitor volcanic activity remotely [4]. Plank et al. [5] used multi-sensor satellite remote sensing data to track volcanic activity at Tonga volcano before its eruption. Similarly, Aiuppa et al. [6] monitored Masaya Volcano, Nicaragua, using gas emissions and Volcanic Radiative Power (VPR) data derived from thermal remote sensing. Their findings showed a strong correlation between VPR data and eruption intensity. Naismith et al. [7] also utilized VPR and other remote sensing data to observe activity at Fuego Volcano. While these studies demonstrate the effectiveness of remote sensing in detecting volcanic activity, they primarily serve as monitoring tools. Remote sensing data provide valuable real-time insights but lack the ability to forecast eruptions or other volcanic activities. Therefore, integrating remote sensing with advanced forecasting techniques, such as machine learning and probabilistic modeling, is essential for improving eruption predictions.

Remote sensing technologies, such as satellite imagery, have significantly improved our ability to monitor volcanic activity remotely and offer a consistent dataset for volcanoes that lack sufficient ground-based sensors [4]. Plank et al. [5] used multi-sensor satellite remote sensing data to track volcanic activity at Tonga volcano before its eruption. Similarly, Aiuppa et al. [6] monitored Masaya Volcano, Nicaragua, using gas emissions and Volcanic Radiative Power (VPR) data derived from thermal remote sensing. Their findings showed a strong correlation between VPR data and eruption intensity. Naismith et al. [7] also utilized VPR and other remote sensing data to observe activity at Fuego Volcano. While these studies demonstrate the effectiveness of remote sensing in detecting volcanic activity, they primarily serve as monitoring tools. Remote sensing data provide valuable real-time insights but so far have lacked the ability to forecast eruptions or other escalations in volcanic activity. Therefore, integrating remote sensing with advanced forecasting techniques, such as machine learning and probabilistic modeling, is essential for improving eruption predictions.

Forecasting volcanic eruptions remains one of the most important goals in volcanology. Historically, seismic approaches have been the primary method for eruption forecasting [8,9]. More recently, machine learning (ML) algorithms have been applied to seismic data to enhance forecasting accuracy [10]. Gaddes et al [11] integrated interferograms derived from radar data with ML techniques to predict volcanic unrest. However, the majority of volcanic forecasting approaches still rely on time series monitoring and trend analysis [5,7].

This research focuses on forecasting volcanic activity using thermal remote sensing data and machine learning. Previous studies have shown that Volcanic Radiative Power (VPR) values fluctuate in response to different eruptive phases. However, accurately forecasting VPR values remains a significant challenge due to the complex and dynamic nature of volcanic systems. In this study, we employ Bayesian Regularized Neural Networks (BRNN) [12] to analyze historical VPR data and predict future thermal activity at Fuego Volcano, aiming to improve short-term eruption forecasting capabilities, where temporal predictors were optimized by relative entropy.

## II. STUDY AREA AND DATA

Fuego Volcano in Guatemala is one of the most active stratovolcanoes in Central America, exhibiting persistent Strombolian activity and periodic explosive eruptions. Since 1524, more than 50 eruptions with a Volcanic Explosivity Index (VEI) of ≥2 have been recorded, making Fuego one of the region's most active volcanoes [13]. Its eruptive history includes both violent Strombolian [14, 15] and sub-Plinian eruptions [16,17].

Fuego is located near the triple junction of the North American, Cocos, and Caribbean tectonic plates (Fig. 1). The complex interaction of compressive and translational forces between these plates influences the behavior of the Central American volcanic Arc [18, 19]. This arc is divided into seven volcanic lineament segments, with Fuego positioned at the northernmost end [20, 21]. Fuego is the most active volcanic center of the Fuego-Acatenango massif, with an estimated upper age limit of 30,000 years [22]. A minimum age of 8,500 years has been inferred by extrapolating effusion rates from a sequence of lavas on Meseta's flank [23].

Thermal data acquired by MODIS sensors on board NASA's Aqua and Terra satellites were analyzed using MIROVA, an automated, near-real-time volcanic hot spot detection method [24]. The MIROVA system uses the mid-infrared (MIR) radiance measured by the two MODIS sensors that acquire multiple daytime and nighttime images of the entire of the entire earth surface with a nominal ground resolution of 1 km every day. These features make MODIS particularly suitable for near-real time monitoring of worldwide volcanic activity [25, 26].

The hot spot detection algorithm consists of contextual spectral and spatial principles specifically designed to efficiently detect small-scale to large-scale thermal anomalies (from <1 MW to >40 GW) while maintaining false detections low. This procedure allows to track a large variety of volcanic activity, including lava flows [27] and domes [28, 29], and strombolian activity [30]. Nonetheless, MIROVA has shown its efficiency in detecting previously unknown effusive activity at remote volcanoes [31], in tracking the development of intense fumarolic activity at Santa Ana volcano ([32] and in detecting the rebirth of Nyamulagira lava lake [24].

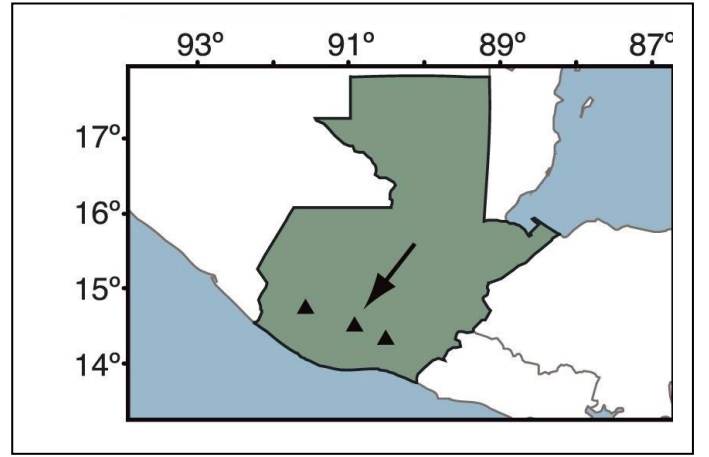

Fig. 1. Fuego Volcano, identified with the arrow, sits between two other active volcanoes in Guatemala named Pacaya and Santiaguito

The Wooster et al. [33] formulation is used to retrieve the volcanic radiative power (VRP; W) from MODIS data, starting from hot spot pixels detected by MIROVA:

$$VPR = 1.89 \times 10^7 \times (L_{MIR} - L_{MIRbk}) \quad (1)$$

where $L_{MIR}$ and $L_{MIRbk}$ are the MIR radiances (W m$^{-1}$ sr$^{-1}$ µm$^{-1}$) characterizing the single hot spot pixel and the background. The constant $1.89 \times 10^7$ (m$^2$ sr µm) is the regression coefficient of the best fitted regression model between above ground MIR radiance and radiant power. The VPR dataset for Fuego Volcano is obtained from the MIROVA system. The dataset spans multiple years and includes daily VPR values. Preprocessing steps include handling missing values, smoothing fluctuations, and normalizing data for model training.

## III. METHODOLOGY

### A. Forecasting model of VPR data

In this paper, we have used a time series model for forecasting the VPR values. The time series model can be treated as a regression problem where the relationship between historical VPR data and future VPR data is achieved by a supervised learning algorithm. The future VPR is then obtained from the developed relationship between the current and past VPR values. For time series modeling, the observed VPR data over time denoted as $Y = [y_1, y_2, ..., y_t]$, where $y_t$ is VPR value at time $t$; is considered as samples from stationary and ergodic process[34].

The time series forecasting for VPR is considered as modeling the relationship of the VPR in time "t" and the VPR values of previous elements of the time series (*t-1*, *t-2*,.., *t-p*) to obtain a function $f$ as it is shown in

$$y_{(p+i+1)} = f(y_{(i+1)}, y_{(i+2)}, ..., y_{(i+p)}), i\epsilon\{1,2,...,t-p-1\} \quad (2)$$

where $y_{(i+1)}, y_{(i+2)}, ..., y_{(i+p)}$ is an input time-lagged VPR values; $y_{(p+i+1)}$ is the forecasted VPR value at time $t = (p + i + 1), i\epsilon\{1,2,...,t-p-1\}$, where $p$ is the number of past time periods VPR values to relate the forecasted VPR value, and function $f$ is non-linear function. Before developing a model to forecast VPR, an initial step, i.e., normalizing the data has to be performed with the observed

VPR values. The original values for $y_{(p+i+1)}$ are normalized to $N_{(p+i+1)}$ within the range [0, 1].

In this paper, a one-step ahead forecast model is developed using a neural network (NN), which requires only one neuron at the output layer; however, multi-step ahead forecasts can also be built by iteratively using one-step ahead predictions as inputs [35]. To develop the NN model for VPR forecasting, the pattern data set extracted in Eq. (2) from the VPR data, i.e., $Y = [y_1, y_2, ..., y_t]$ is normalized.

*B. Input variable selection by relative entropy*

The key task for VPR forecasting is a selection of the number of past time lags, *p*. It is generally selected either by conducting multiple experiments or using statistical methods. Traditional time series methods use correlograms and periodograms for such selection; however, both correlograms and periodograms are based on the assumptions of linear temporal dependency. For measuring nonlinear dependency, [36] develop a statistic based on the mutual information coefficient, known as relative entropy [37]. The relative entropy function δ measures the distance between a joint distribution and the marginal distribution considering the random variables are independent. Relative entropy is calculated by using the Shannon entropy values. The relative entropy δ is calculated as

$$\delta(x, y) = H(x) + H(y) - H(x, y) \quad (3)$$

where $H(x) = -\int_x \log\{f_x(x)\} f_x(x) dx$ for a *p*-dimensional random variable $x$ and $f_x(x)$ is probability distribution function. In this paper, *p*-dimensional random variable $x$ is the input vector $\{y_{(i+1)}, y_{(i+2)}, ..., y_{(i+p)}\}$ in Eq (2). By varying the *p* value, different relative entropy value, δ, can be calculated for the VPR data $Y = [y_1, y_2, ..., y_t]$. High relative entropy represents that the temporal data are independent of each other. Therefore, we have selected the number of past time periods *p*, by observing the trend of the relative entropy values.

The VPR forecasting model was developed using an NN algorithm [38]. This method is capable of achieving pattern recognition by providing an approximate expression for an objective function. The input nodes (variables) for the NN model are $\{N_{(i+1)}, N_{(i+2)}, ..., N_{(i+p)}\}$, and the output node (variable) is $N_{(p+i+1)}$. The number of nodes in the hidden layer was not considered as a predefined value, it was selected by a grid search algorithm. These nodes are interconnected with variable connection strengths. Each node operates by multiplying each incoming signal by a weight and then summing the weighted inputs.

*C. Neural network model for VPR*

The supervised learning step changes these weights in order to reduce the chosen error function, generally mean squared error, in order to optimize the network for use on unknown samples. A major issue for these techniques is the potential for overfitting and overtraining which leads to a fitting of the noise and a loss of generalization of the network. To reduce the potential for overfitting, a mathematical technique known as Bayesian regularization was developed to convert nonlinear systems into "well posed" problems [39, 40]. In general, the training step is aimed at reducing the sum squared error of the model output and target value. Bayesian regularization adds an additional term to this equation:

$$F = \beta E_D + \alpha E_w \quad (4)$$

where F is the objective function, $E_D$ is the sum of squared errors, $E_w$ is the sum of square of the network weights, and α and β are regularization parameters [40]. In the Bayesian network, the weights are considered random variables, thus their density function is written according to the Baye's rules [41]. The optimization of the regularization parameters α and β require solving the Hessian matrix of F(w) at the minimum point $w_{MP}$. Forsee and Hagan [41] proposed a Gauss–Newton approximation to the Hessian matrix which is possible if the Levenburg–Marquardt (LM) training algorithm is used to locate the minimum. This technique reduces the potential for arriving at local minima, thus increasing the generalizability of the network. The main reason we select Bayesian NN is that 1) it reduces overfitting by constraining the parameter space with the prior, and 2) it can handle uncertainty which is often in time series modeling due to the noise in the data.

IV. RESULTS AND DISCUSSIONS

The daily VPR data from April 2000 to December 2019 is used in this study. Non-available samples were removed from the dataset before modeling, resulting in a total of 4,713 data points. Fig. 2 presents the time series of VPR data.

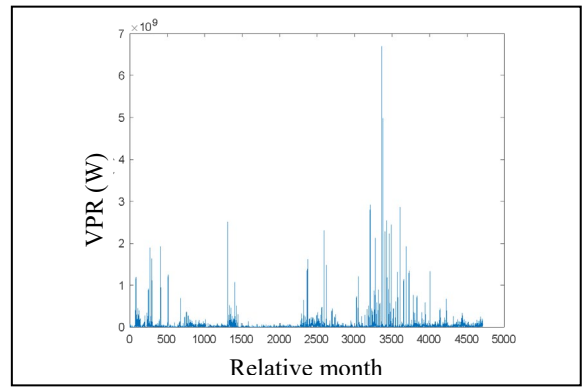

Fig. 2. Historical VPR values (in W) from 2000 to 2019 over Fuego Volcano

Since the time series models assume stationarity, a stationarity test was conducted using the KPSS test (Kwiatkowski et al., 1992). The results rejected the null hypothesis, indicating that the data is non-stationary. Consequently, difference stationarity was tested using the Box et al. [42] method, and the residual values were computed. A second KPSS test was then performed on the residual values, which failed to reject the null hypothesis, confirming that the differenced data is stationary. Therefore, the difference stationary model was adopted, and the residual values were used as input for the neural network model. Fig. 3 presents the residual values from the stationary model.

Before developing the neural network models, input and output patterns were extracted from the residual values of VPR. To determine the optimal number of past time periods (*p*) to use as input variables, relative entropy was calculated for varying p-values. Fig. 4 shows the relative entropy values for different lagged values of *p*, indicating no significant improvement beyond *p* = 6. Based on this, the optimal number of input variables for the neural network model for VPR forecasting was determined to be 5.

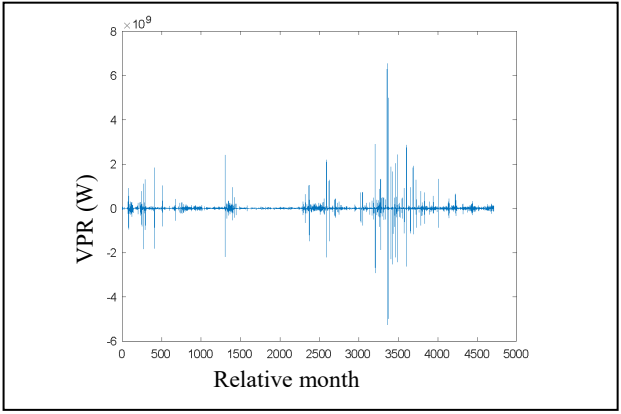

Fig. 3. Residual VPR values (in W) from 2000 to 2019 after taking first difference

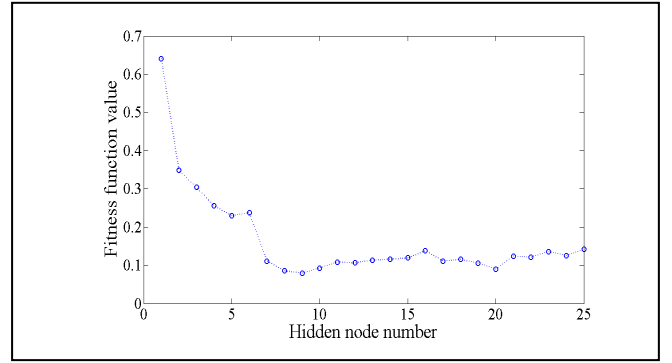

Fig. 5. The objective function value (mean squared error) of BRNN model for different hidden node numbers

TABLE I. TRAINING AND TESTING DATA ERROR STATISTICS OF VPR FORECASTING MODEL

| Statistics | VPR Data | |
|---|---|---|
| | *Training* | *Testing* |
| Mean error (W) | 8.5E-05 | 6.7E-05 |
| Mean squared error (W) | 3.92E+16 | 1.77E+16 |
| $R^2$ | 0.55 | 0.50 |

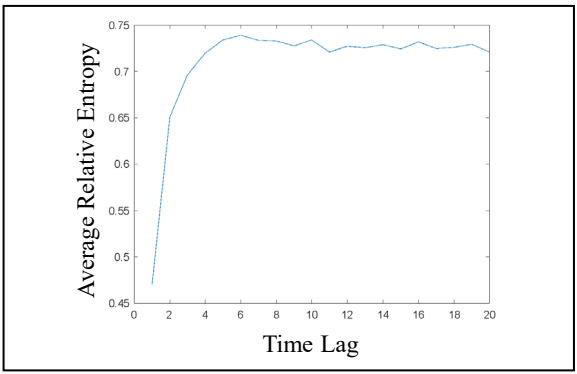

Fig. 4. Average relative entropy value for different time lags. After lag 6, the entropy value was stabilized

A total of 4,706 patterns were extracted from the VPR data for a lag value of 6. Of these, the first 3,765 patterns (approximately 80% of the total) were used as training data for neural network model development, while the remaining 941 patterns (approximately 20%) were reserved for testing. Before modeling, the patterns were normalized to ensure consistency in scale. To verify the statistical similarity between the training and test datasets, a two-sample t-test was performed. The results confirmed that both datasets have the same mean value, indicating no significant difference between them.

The first stage of the experiment involved determining the optimal network size for the BRNN. The number of hidden layer neurons was systematically adjusted in each iteration until the objective function value stabilized. The number of hidden neurons was varied between 2 and 25, with 9 neurons identified as the optimal value, as shown in Fig. 5.

To evaluate the model's performance, error statistics were computed by subtracting the estimated values from the target VPR values, as presented in Table 1. The results indicate that the developed model slightly underestimates VPR for both the training and test datasets. The high mean squared error values (3.92E+16 for the training set and 1.77E+16 for the test set) support the smoothing effect phenomenon, where high-value samples tend to be underestimated, while low-value samples are overestimated. This effect could be significantly reduced by using a simulation algorithm instead of an estimation algorithm; however, such an approach is beyond the scope of this study.

The model's ability to capture data variability is reflected in the $R^2$ value, which represents the proportion of variability explained by the model. In this study, the model captured more than 55% of the data variability (50% for the test dataset), which is reasonably high and indicates that the model fits the VPR dataset well. The close proximity of $R^2$ values for both the training and test datasets suggests that the developed model is well-generalized and can perform consistently with unseen data.

For statistical comparison, a paired-sample t-test was conducted to assess the similarity between actual and predicted VPR values, with the null hypothesis stating that the means of the two populations are different. The results indicate that the null hypothesis can be rejected, confirming that there is no significant difference in means between actual and predicted VPR values for both the training and test datasets.

Additionally, the reproduction of autocorrelation was analyzed by computing the autocorrelation function of actual and forecasted VPR values in the test dataset. Figure 7 presents the autocorrelation values at different lags for both actual and forecasted test data, demonstrating that the autocorrelation values of the forecasted data closely match those of the actual test data.

Finally, a comparative study with other existing methods was performed. We have selected Scaled Conjugate Gradient (SCG) and Levenberg-Marquardt (LM) models along with BRNN. The error statistics (Table 2) indicate that the BRNN outperforms both the SCG and LM models in forecasting VPR values. BRNN achieves the lowest mean error (6.57E-05) and mean squared error (1.77E+16), while also attaining the highest R² value (0.50), meaning it captures 50% of the data variability. In contrast, SCG and LM exhibit higher MSE values (4.1E+16 and 5.5E+16, respectively) and lower R² scores (0.22 and 0.14), suggesting they strive to generalize well to the dataset. The improved performance of BRNN can be attributed to its Bayesian regularization, which mitigates overfitting and enhances generalization. While BRNN provides the most accurate predictions, its R² value indicates that further refinements, such as incorporating additional input features or optimizing hyperparameters, could enhance model performance.

TABLE II. FORECASTING ERROR STATISTICS OF DIFFERENT NEURAL NETWORK MODELS WITH DIFFERENT NEURAL TRAINING ALGORITHMS, INCLUDING BRNN

| Statistics | Different Neural Network training algorithms | | |
|---|---|---|---|
| | *SCG* | *LM* | *BRNN* |
| Mean error (W) | 8.37E-05 | 7.52E-05 | 6.7E-05 |
| Mean squared error (W) | 4.1E+16 | 5.5R+16 | 1.77E+16 |
| R² | 0.22 | 0.14 | 0.50 |

## V. CONCLUSIONS

This study demonstrates the effectiveness of BRNN for forecasting VPR based on historical thermal remote sensing data at Fuego Volcano. This research adopted an entropy-based approach to select optimum lag value for the forecasting model. Compared to SCG and LM training models, BRNN achieved the lowest mean error and mean squared error while capturing the highest proportion of data variability. These results highlight the advantages of Bayesian regularization in improving model generalization and reducing overfitting. The findings suggest that machine learning, particularly BRNN, can serve as a valuable tool for analyzing complex volcanic activity patterns and supporting eruption monitoring efforts. However, while BRNN performed the best among tested models, the moderate R² value indicates that additional factors influence VPR fluctuations, necessitating further refinement of forecasting approaches.

Future research should focus on enhancing predictive accuracy by integrating additional geophysical indicators, such as seismic activity, gas emissions, and deformation data, into the forecasting models. Exploring hybrid approaches that combine deep learning techniques, such as Long Short-Term Memory (LSTM) networks, with traditional statistical models could improve temporal pattern recognition and long-term forecasting reliability. Additionally, incorporating uncertainty quantification methods would provide confidence intervals for predictions, increasing their reliability for hazard assessment. Further validation using real-time volcanic activity data from different volcanic systems would also help generalize the model for broader applications in volcanic hazard forecasting.

ACKNOWLEDGMENT

We acknowledge the MIROVA system for providing VPR data and the scientific community for their contributions to volcanic monitoring and modeling